\documentclass[runningheads]{llncs}
\usepackage{graphicx}
\usepackage{xcolor}
\usepackage{enumitem}
\usepackage{booktabs}
\begin{document}
\title{EPARS: Early Prediction of At-risk Students with Online and Offline Learning Behaviors}
\titlerunning{EPARS}
%
\author{Yu Yang \inst{1,*} \and
	Zhiyuan Wen \inst{1,*} \and
	Jiannong Cao \inst{1} \and
	Jiaxing Shen \inst{1} \and
	Hongzhi Yin \inst{2} \and
	Xiaofang Zhou \inst{2}
}

\institute{Department of Computing, The Hong Kong Polytechnic University \\
	\email{\{csyyang,cszwen,csjshen\}@comp.polyu.edu.hk} \\
	\email{jiannong.cao@polyu.edu.hk} \\
	\and School of ITEE, The University of Queensland, \\
	\email{h.yin1@uq.edu.au} \\
	\email{zxf@itee.uq.edu.au}}

\footnotetext[1]{These authors have contributed equally to this work.}

\authorrunning{Y. Yang et al.}
%
%
\maketitle              
\begin{abstract}
Early prediction of students at risk (STAR) is an effective and significant means to provide timely intervention for dropout and suicide.
Existing works mostly rely on either online or offline learning behaviors which are not comprehensive enough to capture the whole learning processes and lead to unsatisfying prediction performance.
We propose a novel algorithm (EPARS) that could early predict STAR in a semester by modeling online and offline learning behaviors.
The online behaviors come from the log of activities when students use the online learning management system.
The offline behaviors derive from the check-in records of the library.
Our main observations are two folds.
Significantly different from good students, STAR barely have regular and clear study routines.
We devised a multi-scale bag-of-regularity method to extract the regularity of learning behaviors that is robust to sparse data.
Second, friends of STAR are more likely to be at risk.
We constructed a co-occurrence network to approximate the underlying social network and encode the social homophily as features through network embedding.
To validate the proposed algorithm, extensive experiments have been conducted among an Asian university with $15,503$ undergraduate students.
The results indicate EPARS outperforms baselines by $14.62\% \sim 38.22\%$ in predicting STAR.


\end{abstract}
\section{Introduction}

Predicting students at risk (STAR) plays a crucial and significant role in education as STAR keep raising public concern of dropout and suicide among adolescents \cite{orozco2018association,stinebrickner2014academic}.
STAR refer to students requiring temporary or ongoing intervention to succeed academically \cite{richardson2005risk}.
Students may be at risk for several reasons like family problems and personal issues including poor academic performance.
Those students will gradually fail to sustain their studies and then drop out which is also a waste of educational resources \cite{berens2018early}. 
Early prediction of STAR offer educators the opportunity to intervene in a timely manner.

Traditionally, many universities identify STAR by their academic performance which sometimes is too late to intervene.
Existing works are largely based on either online behaviors or offline behaviors of students \cite{he2015identifying,koprinska2015students,MARBOUTI20161}.
For example, STAR are predicted in a particular course from in-class feedback such as the grade of homework, quiz, and mid-term examination \cite{MARBOUTI20161}. 
However, due to the complex nature of STAR \cite{ellenbogen1997peer}, either online and offline behaviors only capture part of the learning processes.
For example, some students prefer learning with printed documents so they become inactive in online learning platforms after downloading learning materials. This process is difficult to capture through their online learning behaviors. 
Therefore, existing work can hardly capture the whole learning processes in a comprehensive way and thus leads to poor performance in the early prediction of STAR.

In this work, we aim to predict STAR before the end of a semester using both online and offline learning behaviors. 
STAR are defined as students with an average GPA below $2.0$ in a semester. 
Online behaviors are extracted from click-stream traces on a learning management system (LMS).
These traces reveal how students use various functionalities of LMS.
While the offline behaviors derive from library check-in records.
To achieve the goal, we encounter the following three major challenges: (1) \textbf{Lable imbalance}. 
The number of STAR is significantly smaller than that of normal students, which makes it an extreme label-imbalance classification problem. The classifier will be easily dominated by the majority class (normal students). 
(2) \textbf{Data density imbalance}. The library check-in records are much sparser than click-stream traces on the online learning platform so that it is challenging to fuse them fairly well for classifying STAR. 
(3) \textbf{Data insufficiency}. Students, especially STAR, are usually inactive at the early stage of a semester. As a result, the behavior traces are far from enough for accurate early prediction of STAR.

In light of these challenges, we propose a novel algorithm ({EPARS}) for \underline{e}arly \underline{p}rediction of \underline{a}t-\underline{r}isk \underline{s}tudents. 
EPARS captures students' regularity patterns of learning processes in a robust manner.
Besides, it also models social homophily among students to perform highly accurate early STAR prediction. 
The intuitions behind EPARS are two-fold. 
First, good students usually follow their study routines periodically and show clear regularities of learning patterns \cite{yao2019predicting}. However, the study routines of STAR are disorganized leading to irregular learning patterns, which is different from good students. 
Second, students tend to have social tie with others who are similar to them according to the theory of social homophily \cite{marsden1988homogeneity} and existing studies found that at-risk students had more dropout friends \cite{ellenbogen1997peer}.

Based on both intuitions, we first propose a multi-scale bag-of-regularity method to extract discriminative features from the regularity patterns of students' learning behaviors.
Unlike the traditional approaches using entropy for measuring the regularities, which cannot work well on sparse data, we ignore the inactive behavior subsequence and capture the regularity patterns in a multi-scale manner. 
Our approach can capture the regularity patterns fairly well even though the data are very sparse.
Therefore, it overcomes the challenge of data density imbalance and extracts discriminative features from regularity patterns for classifying STAR.
In order to model the social homophily, we construct a co-occurrence network from the library check-in records to approximate social relationships among students.
Co-occurrence networks have been widely used in modeling social relationship and achieved great success in many application scenarios \cite{shen2019bag,shen2018snow}
After that, we embed the co-occurrence networks and learn a representation vector for every student with the assumption that students' representation vectors are close when they have similar social connections.
Modeling the social homophily provides extra information to supplement the lack of behavior trace for STAR at the beginning of a semester, which solves the data insufficiency problems and makes EPARS capable of early predicting STAR.
Moreover, we oversample the training samples of STAR by random interpolating using SMOTE \cite{chawla2002smote}, which overcomes the label imbalance problem while training the classifiers.

We conducted extensive experiments on a large scale dataset covering all $15,503$ undergraduate students from freshmen to senior students in the whole university.
The experimental results show that the proposed EPARS achieves $0.7237$ accuracy in predicting STAR before the end of a semester and $0.6184$ prediction accuracy after the first week of the semester, which outperforms the baseline by $34.14\%$ and $38.22\%$ respectively.
Comparative experiments found that our proposed multi-scale bag-of-regularity method and modeling students' social homophily by the co-occurrence network improve the performance of STAR early prediction $26.82\%$ and $14.62\%$ respectively.
From the data analysis, we also found that STAR engaged less than normal students in learning in the early semester.
Besides, the results confirm that the friends of STAR are more likely to be at risk if they have similar regularity patterns of learning behaviors, which in line with the conclusion drawn by an existing experimental study \cite{ellenbogen1997peer}.

The our contributions are summarized as follows.
\begin{itemize}
	\item We propose a multi-scale bag-of-regularity approach to extract regularity patterns of learning behaviors, which is robust for sparse data. This approach is also generic for extracting repeated patterns from any given sequence.
	\item We model the social homophily among students by embedding a co-occurrence network constructed from their library check-in records, which reliefs the data insufficiency issues. 
	\item Extensive experiments on a university-scale dataset show that our proposed EPARS is effective on STAR early prediction in terms of $14.62\%\sim38.22\%$ accuracy improvement to the baselines.
\end{itemize}
	
The remainder of this paper is organized as follows. We review the relative works in the next section and formally formulate the STAR early prediction problem in section 3. The data description are reported in section 4. In section 5, we present the proposed EPARS in detail and evaluate its effectiveness in section 6 before we conclude the paper in the last section.

\section{Related Works}

There are various reasons for students being at-risk, including school factors, community factors, and family factors.
Most of the existing works focus on school factors due to the convenience of data collection.
The classification models used include Logistic Regression, Decision Trees, and Support Vector Machines.
The main difference of these works relies on the input features, which could be generally classified into offline and online. 

The offline learning behaviors contain check-ins of classes or libraries, quiz and homework grades, and records of other activities conduct in the offline environment.
These kinds of works are quite straight forward to monitor the student learning activities for identification. 
Early researchers design the Personal Response system and utilize the order of students'  device registration to help identify STAR \cite{griff2008early}.
Besides, questionnaires and personal interviews are also applied to collect student information for identification \cite{choi2018learning}.
These methods show accurate results in an early stage of a semester.
Moreover,  Marbouti et al. also proposed to identify STAR at three time-points (week 2, 4, and 9) in a semester using in-term performance consists of homework and quiz grades and mid-term exam scores \cite{MARBOUTI20161}. 
These methods rely heavily on domain knowledge, and collecting these offline learning data is very high labor cost and time-consuming, such that they are not practical for large scale STAR prediction.

With the popularization of online learning, researchers have turned their attention to analyzing student behavioral data on online learning platforms such as MOOCs and Open edX. The online learning behaviors are collected from the trace that students left in the online learning system such as click-stream logs in functional modules of the systems, forum posts, assignment submission, etc. 
Kondo et al. early detect STAR from the system login and assignment submission logs on the LMS \cite{kondo2017early}, but their results may be partial since most students are not actively engaged with LMS.
Shelton et al. designed a multi-tasks model to predict outstanding students and STAR \cite{shelton2018two}, which purely uses the frequency of module access as features.
\cite{ho2018data} proposed a personalized model for predicting STAR enrolling in different courses, but it is hardly generalized to various courses, especially the totally new one.
Instead of purely using statistic features, we further extract students' regularity patterns and social homophily for early predicting STAR.

\section{Problem Formulation}
This section gives the formal problem definition of STAR early prediction which is essentially a binary classification problem.
We will introduce the exact definition of STAR, the input data, and the meaning of early prediction.

According to the student handbook of the university, when a student has a Grade Point Average (GPA) lower than $2.0$, he/she will be put on academic probation in the following semester.
If a student is able to pull his/her GPA up to $2.0$ or above at the end of the semester, the status of academic probation will be lifted.
Otherwise, he/she will be dropped out.
Therefore, we define STAR as students whose average GPA is below $2.0$ in a semester. 

The input data are two folds. 
One is the records of students' online activities in the Blackboard, a learning management system.
The Blackboard has several modules including course participation, communication and collaboration, assessment and assignments. 
Students could browse and download course-related materials including lecture keynotes, assignments, quizzes, lab documents etc. 
They can also take online quizzes and upload their answers for assessment. 
Besides, students could communicate over the different posts and collaborate on their group assignments.
Students' click operations in the Blackboard will be recorded (online traces).
The other is the check-in records of the library. 
Students have to tap their student cards before entering the library (offline records).

Early prediction means the input data are collected before the end of a semester. 
Given online traces and offline records accumulated within $t$ $(t<t_{end})$ where $t_{end}$ is the end time of a semester, our objective is to identify STAR as accurate as possible.

\section{Data Description}
We collect students' online and offline learning traces and their average GPA in an Asian University in 2016 to 2017 academic year. The online learning traces come from how students use the Blackboard, a learning management system, to learn. There are many functions in the Blackboard but some of them are rare to be used by students. Thus, we collect the click-stream data with timestamps from some of the most popular modules in the Blackboard including log-in, log-out, course materials access, assignment, grade center, discussion board, announcement board, group activity, personal information pages, etc. Offline learning traces come from students' library check-in records which indicating when they go to library. Since students do not need to tap their student cards when they leave the library, the check-out records will not be marked down and we exclude it in this study.

\begin{table}
\centering
\caption{Data Overview.}
\begin{tabular}{l|cc|cc}   
\toprule
& \multicolumn{2}{c|}{Semester 1} & \multicolumn{2}{c}{Semester 2}\\
\cline{2-5}
& STAR & Other Std & STAR & Other Std \\
\hline
Population & 391 & 15,112 & 225 & 15,278 \\
\# click-stream logs in LMS & 2,225,605 & 95,949,014 & 1,019,134 & 70,874,428 \\
Avg. \# click-stream logs & 5,692.0844 & 6,349.1936 & 4,529.4844 & 4,638.9860 \\
Avg. \# click-stream logs in first 2 weeks & 301.4041 & 399.9502 & 243.0400 & 284.4368 \\
Avg. \# click-stream logs in last 2 weeks & 526.6522 & 545.4346 & 336.9133 & 304.7331 \\
\# library check-in & 14,045 & 636,353 & 6,245 & 517,557 \\
Avg. \# library check-in & 35.9207 & 42.1091 & 27.7556 & 33.8760 \\
Avg. \# library check-in in first 2 weeks & 1.7877 & 2.3303 & 1.3889 & 1.8424 \\
Avg. \# library check-in in last 2 weeks & 2.9834 & 3.3760 & 2.3444 & 2.4547 \\
\bottomrule
\end{tabular}
\label{Tab:Data_overview}
\end{table}
	
All $15,503$ undergraduate students in the whole university involved in this study. Every student has a unique but encrypted ID for linking their LMS click-stream data, library check-in records, and GPA. The overview of collected data are showed in Tab. \ref{Tab:Data_overview}. There are $225$ and $319$ STAR in semester one and two respectively, which are $1.45\%$ and $2.06\%$ of all students. This makes our STAR early prediction as an extremely label imbalance classification problem, which is our first challenge. In addition, students left over $170$ million click-stream logs but only $1.7$ million library check-in records in the whole academic year such that the data density between online and offline learning trace are also imbalance. Compared to the last two weeks of the semester, all students are less active in the first two weeks and STAR are even less active than normal students which cause data inefficiency problems for early predict STAR at the beginning of the semester.

\section{Methodologies}
In this section, we will elaborate on the proposed EPARS including multi-scale bag-of-regularity, social homophily, and data augmentation.

\subsection{Multi-scale Bag-of-Regularity}
In order to extract the regularity patterns from students' learning traces, we propose multi-scale bag-of-regularity here, which is robust for sparse data.
	
Based on Hugh Drummond's definition, behavior regularity is repeatedly occurring of a certain behavior in descriptions of patterns \cite{drummond1981nature}. Students usually have their own repeated patterns for using LMS and going to the library. For instance, some students prefer to go to the library every Monday and Thursday. It is possible for us to illustrate their repeated patterns on multiple scales such as they will not go to library after the day they go there; they go to the library two and three days apart alternately. If we purely extract the regularity patterns on a single scale, it hardly captures the complete picture and leads to information loss. This motivates us to extract the regularity patterns in multi-scales. In addition, traditional approaches, such as entropy, measure the regularities in a global perspective. When students' library check-in data are sparse, those approaches will regard their library check-in as outliers and consider their general regularity patterns as never go to the library, which are incorrect. Therefore, we focus on the every behavior trace students leave during learning for extracting their learning regularity patterns.
	
First of all, we construct a binary sequence from students' behavior traces. When they have certain behaviors, such as check-in to the library, we mark it as $1$ in the sequence. The time granularity for constructing the binary sequence depends on the application and the time granularity we used in this study is a day. Next, We sample subsequences of length $\ell$ centered on every nonzero element in the sequence. The length of subsequences $\ell=2+(s-1)\times z$ where $s\in\{1,2,\cdots,S\}$ is scale and $z$ is the step-size between scales. This sampling approach guarantees that no all-zeros sequence will be sampled for the following regularity measurement which gives our method the ability to overcome data sparsity issues. Every subsequence actually is a behavior pattern that is viewed on different scales.
	
After sampling the behavior patterns, we explore the repeated patterns from them to obtain the regularities. Since the regularity is repeatedly occurring of behavior patterns, we ignore the subsequences that the times of occurrences are less than a threshold $n$. For the subsequence of length $\ell$ in scale $s$, it contains $2^\ell-1$ different behavior pattern excluding all-zeros one. We regard them as a bag and count the number of occurrences of every behavior pattern. Finally, a $(2^\ell-1) \times 1$ vector $r_s$ is obtained, which carries the behavior regularities on scale $s$. Lastly, we concatenate the regularity vectors $r_s$ in every scale as the representation of regularity on multi-scales. Our bag-of-regularity approach explores the regularity patterns of behaviors in multi-scales such that it can extract richer information from the sparse input sequence. The regularity features extracted from dense LMS data and sparse library check-in records by our multi-scale bag-of-regularity are on the same scale-space so that we can simply concatenate them together as the final regularity features for STAR prediction and the performance is fairly well. In addition, the proposed multi-scale bag-of-regularity is generic for extracting repeated patterns from any given sequence since it will transform the input sequence into a binary sequence before extracting regularities.

\subsection{Social Homophily}
We construct a co-occurrence network to model the social relationship among students. If students are friends, they are more likely to learn together because of the social homophily \cite{marsden1988homogeneity}. They have a higher probability to go to the library together comparing to strangers. Thus, we assume that two students are friends if they go to library together. If the time difference of the library check-in between two students is less than a threshold $\delta$, we treat this as the co-occurrence of two students in the library. In other words, they go to the library together. Based on this, we construct a co-occurrence network $G(V,E,W)$ where nodes $V$ are students and there is an edge $e\in E$ linking two nodes if students go to the library together. Each edge is accompanied by a weight value $w\in W$ showing how many times they co-occurrence in the library. We constrain $w\geq \sigma$ which is a threshold to filter out the ``familiar strangers". We do not construct the co-occurrence network from the LMS log-in traces because the LMS log-in frequency is too high and it will involve too many ``familiar strangers" in the network. This will introduce significant biases for learning the social homophily later.

Next step is to learn students' social homophily from the co-occurrence network.
Network embedding has been widely applied in encoding the connectivities among nodes as representation and well preserves the graph properties \cite{li2017ppne,yang2017properties}. 
Here, we embed the co-occurrence network by Node2Vec \cite{grover2016node2vec} and learn a representation vector for every node which preserves the connectivities among students. In addition, we constrain that the learned representation of nodes should be close when they have similar connections. Specifically, we first exploring diverse neighborhoods for every node by a biased random walk. Let us denote $c_i$ as the $i$th node in the walk. We sample node sequences with transition probability
\begin{equation}
	p(c_i=u|c_{i-1}=v)=\left\{ 
		\begin{array}{cl}
			\frac{\alpha_{pq}w_{uv}}{Z} \ \ & \mathrm{if} \ (u,v)\in E\\
			0 \ \ & \mathrm{Otherwise}
		\end{array}	
	\right.
\end{equation}
where $Z$ is a constant for normalization and $\alpha_{pq}$ in Eq. (\ref{Eq: alpha}) is the sampling bias.
\begin{equation}
    \alpha_{pq}=\left\{ 
	    \begin{array}{cl}
	        1/p\ \ & \mathrm{if} \  d_{uv}=0\\
	        1\ \   & \mathrm{if} \ d_{uv}=1\\
	        1/q\ \ & \mathrm{if} \ d_{uv}=2\\
	        0 \ \ & \mathrm{Otherwise} 
	    \end{array}	
    \right.
\label{Eq: alpha}
\end{equation}
$d_{uv}$ denotes the shortest path distance between nodes $u$ and $v$. Parameters $p$ and $q$ make the trade-offs between depth-first and breadth-first neighborhood sampling.

To learning the final representation of every node, we train a Skip-gram model \cite{perozzi2014deepwalk} by maximizing the log-probability of its network neighborhood conditioned on its feature representation as showed in Eq. (\ref{Eq: loss_node2vec}) where $f(\cdot)$ is a mapping function from node to feature representations and $N_s(u)$ is $u$'s neighborhood sampling by the above random walk.
\begin{equation}
	\max_{f}\sum_{u\in V}\log\left(\prod_{v_i \in N_s(u)}\frac{\exp\left(f(u)\cdot f(v_i)\right)}{\sum_{v \in V}\exp\left(f(u)\cdot f(v)\right)}\right)
\label{Eq: loss_node2vec}
\end{equation}
We adopt the stochastic gradient ascent to optimize the above objective function over the model parameters and obtain the representation of every node which carrying its social homophily. Learning students' social homophily provides extra information for dealing with the data insufficiency issues such that it makes our EPARS have the ability to early predict STAR.

\subsection{Data Augmentation}

To deal with the extremely label imbalance issues, we oversample the STAR by a synthetic minority over-sampling technique (SMOTE) \cite{chawla2002smote} while constructing the training set. For each STAR training sample, denoted as $x$, we first search its $k$-nearest neighbors from all STAR samples in training set by the Euclidean distance in the feature space, and the $k$ is set to $10$ in our experiment. Next, we randomly select a sample $x'$ from the $k$ nearest neighbors and synthesize a new STAR example by Eq. (\ref{Eq: Smote}) where $\omega$ is a random number between $0$ and $1$.     
\begin{equation}
    x_{new}=x+(x'-x)\times\omega
\label{Eq: Smote}
\end{equation}

After the data augmentation, STAR have the same amount as the normal students in the training set; this allows the classifier to avoid being dominated by the majority of the normal students during training.
SMOTE synthesizes new examples between any of the two existing minority samples by a linear interpolation approach. Compared with a widely used under-sampling technique EasyEnsemble, SMOTE introduces random perturbation into the training set while generating the synthetic examples, which provide the trained classifier better generalization. 

\section{Experiments}
We conduct experiments to showcase the effectiveness of proposed EPARS. In particular, we aim to answer the following research questions (RQ) via experiments:    
\begin{itemize}
    \item RQ1: How effective is the EPARS in predicting STAR?
    \item RQ2: How early does the EPARS well predict STAR?
    \item RQ3: How effective is SMOTE for data augmentation in EPARS?
    \item RQ4: Is the EPARS sensitive to hyper-parameters?
\end{itemize}
	
\subsection{Experiment Protocol}
\subsubsection{Experiment Setting.}
In our dataset, each student has an independent label of either STAR or the normal student in each semester. 
Thus, we treat students in different semesters as a whole in our experiments.
When predicting STAR at any time $t$ before the end of the semester $t_{end}$, we extract features from their online and offline learning traces from the beginning of a semester to the current time $t$.  After feature extraction, we synthesize new STAR examples to augment the training set. We conduct experiments under the 5-fold cross-validation setting and repeat $10$ times. The average results will be reported in the next subsection. Several classifiers are tested, including the Logistic Regression, Support Vector Machine (SVM), Decision Tree, Random Forest, and the Gradient Boosting Decision Tree (GBDT). GBDT outperforms all other classifiers in our experiments, so we only report the results of GBDT due to the space limit.	

\subsubsection{Parameter Setting.} We set the maximum scale of regularity $S = 4$, the co-occurrence threshold $\delta$ to be $30$ seconds, the linking threshold $\sigma = 2$, and the dimension of embedding to be $64$ for EPARS. We select $k=10$ neighborhood for SMOTE to augment the training set. The classifier GBDT is trained with parameters that the number of estimators is $100$, maximum depth of the decision tree is $10$, and the learning rate is $0.1$.

\subsubsection{Evaluation Metrics.}
We evaluate the performance of EPARS from two aspects. Since the STAR prediction is a binary classification problem, we adopt Area Under the receiver operating characteristics Curve (AUC) to measure the classification performance. The AUC indicates how capable the model is to distinguish between STAR and the normal students. Moreover, since our focus is to find out the STAR as accurate as possible, we measure the accuracy of our model in predicting STAR by the number of true positive predictions divided by the total number of STAR in the test set. We denote it as ACC-STAR, which indicates how many percentages of STAR are correctly predicted. 


\subsubsection{Baseline Approaches.} 
As mentioned in the introduction, our major contribution is to achieve better STAR early prediction performance, in terms of higher AUC and ACC-STAR, with features extracted from students' learning regularity and social homophily. To verify the effectiveness of EPARS, we set four baseline models, including SF, DA, DA-Reg, and DA-SoH. SF uses only the statistically significant behavior features as input to predict STAR without data augmentation. The process of discovering significant statistical features will be presented in the next paragraph. DA uses the same features as SF and augments the training set using SMOTE. Comparing SF and DA, we can verify whether SMOTE can solve the label imbalance challenge well and results in better classification performance. 
DA-Reg and DA-SoH integrate the regularity features and the social homophily to the DA, respectively. They are to verify the effectiveness of our proposed multi-scale bag-of-regularity and the social homophily modeling approach in STAR prediction.

To discover the significant statistical features, we perform an ANOVA (analysis of variance) test to figure out what behaviors are statistically significant for distinguishing between STAR and the normal students. 
We have 13 kinds of clickstream behaviors on the LMS and 28 kinds of library check-in behaviors at different times of the day and different periods in the semester. Due to the space limited, we report the statistically significant features and some of the insignificant features discovered from the ANOVA in Table \ref{Tab: ANOVA}. 
It is interesting to note that STAR use the LMS less than the normal students, but they will check the announcement and lectures' information more. 
There is no significant difference in accessing the course materials and checking assignment results. 
Besides, STAR go to the library less than the normal students at the beginning of a semester. Still, they prefer more to be there after business hours. 
Lastly, we select the statistically significant features as the SF baseline to benchmark our proposed EPARS.


\begin{table}[ht]
	\centering
	\caption{Results of the ANOVA test.}
	\begin{tabular}{l|c|c|c|c}   
		\toprule
		Features & P-value & F-value & Mean STAR & Mean Others\\
		\hline
		\# LMS Login & 0.0020 & 9.5112 & 127.4987 & 144.8043 \\
		\# LMS Logout & 0.0000 & 34.5301 & 8.9318 & 20.1348 \\
		\# Check announcement & 0.0158 & 5.8311 & 41.4436 & 36.8361 \\
		\# Course access & 0.7328 & 0.1165 & 4.2677 & 4.5667 \\
		\# Grade center access & 0.7694 & 0.0859 & 10.5486 & 10.2108 \\
		\# Discussion board access & 0.0020 & 9.5951 & 11.7979 & 19.2444 \\
		\# Group access & 0.0209 & 5.3385 & 13.2782 & 20.1268 \\
		\# Check personal info & 0.0000 & 16.7953 & 0.2283 & 1.6585 \\
		\# Check lecturer info & 0.0000	& 106.1638 & 9.7297 & 5.5440 \\
		\# Journal page access & 0.0199 & 5.4191 & 0.2283 & 1.6585 \\
		\# Lib check-in & 0.0700 & 3.2829 & 42.8163 & 47.3589 \\
		\# Lib check-in in the morning & 0.0001 & 14.7133 & 7.0367 & 9.4206 \\
		\# Lib check-in in the afternoon & 0.0023 & 9.3196 & 27.0604 & 31.9419 \\
		\# Lib check-in after midnight & 0.0000 & 43.9327 & 4.0105 & 1.6927 \\
		\# Lib check-in before exam months & 0.0123 & 6.2740 & 33.9265 & 39.0143 \\
		\# Lib check-in at the first month & 0.0004 & 12.5447 & 8.4724 & 10.6052 \\
	\bottomrule
	\end{tabular}
\label{Tab: ANOVA}
\end{table}

\subsection{Experimental Results}
\subsubsection{RQ1:} To verify the effectiveness of our proposed EPARS in predicting STAR, we extract features from the whole semester data to train the GBDT and benchmark EPARS with four baselines. This experiment evaluates the performance of EPARS when students' all learning behaviors in a whole semester is known. The results are presented in Tab. \ref{Tab: Pred_STAR_result}. 
	
\begin{table}
	\centering
	\caption{Results of predicting STAR using the whole semester learning behavior data.}
	\begin{tabular}{c|c|c|c|c|c}   
		\toprule
		Metric & SF & DA & DA-Reg & DA-SoH & EPARS\\
		\hline
		AUC & 0.8423 & 0.8442 & 0.8611 & 0.8623 & \textbf{0.8684} \\
		ACC-STAR & 0.5395 & 0.6079 & 0.6842 & 0.6184 & \textbf{0.7237} \\
		\bottomrule
	\end{tabular}
	\label{Tab: Pred_STAR_result}
\end{table}

Comparing the experimental results between SF and DA, it is confirmed that our data augmentation approach overcomes the data imbalance challenges to some extent and achieves improvement in both AUC and ACC-STAR. In addition, the regularity features extracted by our multi-scale bag-of-regularity method can improve the accuracy of predicting STAR a lot, which indicates that the regularity of learning is a distinguished feature between STAR and the normal students, and the multi-scale bag-of-regularity can well extract their regularity patterns efficiently. Compared with DA-Reg, DA-SoH achieves a higher AUC score and has better overall classification performance. However, its ACC-STAR is much lower than DA-Reg's, suggesting that it cannot identify STAR as accurate as DA-Reg. In other words, social homophily helps identify the normal students a lot rather than recognizing STAR. 
This shows that our approach is capable of well modeling the social homophily among students. Nevertheless, STAR may have similar linkage patterns with ``familiar strangers" in the co-occurrence network since STAR are very handful. 
Combining the regularity patterns of learning and social homophily, which is our proposed EPARS, achieves the best performance in predicting STA in terms of $19.05\%$, $5.77\%$ and $17.03\%$ ACC-STAR improvement to DA, DA-Reg and DA-SoH, respectively. This indicates that friends of STAR are more likely to be at-risk if their regularity patterns of learning behaviors are also similar. Therefore, the regularity features can help eliminate the ``familiar strangers" and result in better STAR prediction performance.

\subsubsection{RQ2:} 
\begin{figure}[h!]
	\centering
	\includegraphics[scale=0.35]{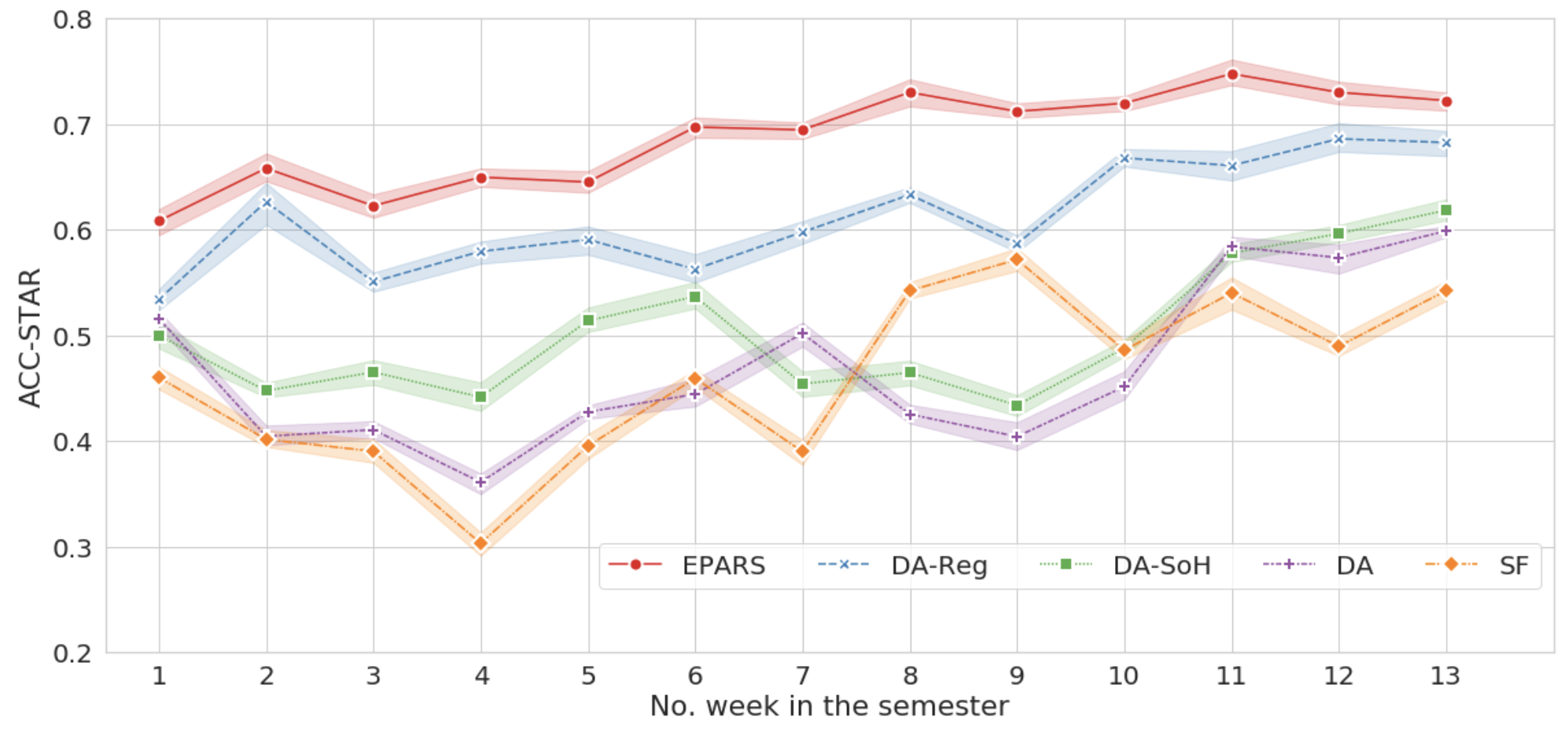}
	\caption{Results of STAR early prediction.}
	\label{fig:rq2}
\end{figure}

To demonstrate the effectiveness of our methods in early predicting STAR, we conduct experiments in every week's data of the semester. For each week, we extract features of students' learning traces from the beginning of the semester to the end of that week. We repeat the experiment for $10$ times, and the average ACC-STAR of early predicting STAR is presented in Fig. (\ref{fig:rq2}) in which the solid lines are the average ACC-STAR, and the shadows represent the error spans.
	
Our EPARS outperforms all other baselines from the first week to the end of the semester. It is worth mention that our EPARS can correctly predict $61.84\%$ STAR only based on the online and offline learning traces of the students in the first week, which outperforms SF, DA, DA-Reg, and DA-SoH $38.22\%$, $17.50\%$, $14.62\%$, and $22.38\%$, respectively. In the first four weeks, the prediction performance of SF keeps on decreasing. One possible reason is that some normal students are not active in the beginning of the semester, so that they may have similar behavior patterns with STAR and cause misclassification. Students' social homophily and regularity patterns of learning behaviors are much more discriminable especially in the early stage of a semester. The performance of EPARS is almost converged in the middle of a semester while other baselines are still gradually increasing or concussion. It shows that our EPARS can leverage less information but achieves better performance in early predicting STAR.

\subsubsection{RQ3:} To verify the effectiveness of using SMOTE for dealing with the label imbalance issues, we conduct a comparative experiment among random undersampling (RU), random oversampling (RO) and SMOTE. RU and RO are widely adopted in existing work for STAR prediction \cite{he2015identifying,jayaprakash2014early}. RU randomly deletes examples with the majority labels until the labels of training samples are balanced while RO randomly resamples the minority examples until the numbers of the minority are the same as the majority one. We regard SF as baseline and launch above data augmentation approach for predicting STAR before the end of a semester. We repeat the experiment 10 times and report the average AUC and ACC-STAR in Tab. \ref{Tab:result_DA}. 

The first two columns show the number of examples in the training set after data augmentation in each fold of the experiment. Experimental results show that RO slightly outperforms the baselines but the performance of RU is worse than the baselines. In the case of extremely label imbalance, undersampling technique drops most of negative training samples and constructs a very small training set, which cannot provide enough information to well train a classifier. Although RO augments the minority examples by oversampling, most synthesis examples are the same so that the classifier is very easy to overfit and results in poor testing accuracy. SMOTE synthesizes the minority examples by linear interpolation which not only increases the number of minority samples but also enriches the diversity of the training set. Thus, it achieves the best STAR prediction accuracy in such an extremely label imbalance classification task.

\begin{table}[h]
	\centering
	\caption{Evaluation of data augmentation.}
	\begin{tabular}{c|c|c|c|c}   
		\toprule
		 & \# STAR after DA & \# Normal Std after DA & AUC & ACC-STAR\\
		\hline
		SF &305& 11295 & 0.8342  & 0.5526    \\			
		RU & 305 & 305 & 0.8211 & 0.5316    \\
		RO & 11295 & 11295 & 0.8458  & 0.5645  \\
		SMOTE & 11295 & 11295 & \textbf{0.8684}  & \textbf{0.7237}  \\
		\bottomrule
	\end{tabular}
	\label{Tab:result_DA}
\end{table}

\subsubsection{RQ4:} We test how sensitive EPARS is to the hyper-parameters and discuss how to select hyper-parameters for EPARS. We focus on three hyper-parameters of EPARS. One is the maximum scale $S$ of multi-scale bag-of-regularity. The other two are co-occurrence threshold $\delta$ and linking threshold $\sigma$ between pairs of students when constructing co-occurrence networks for further modeling the social homophily.

\begin{figure}[h!]
	\centering
	\includegraphics[scale=0.36]{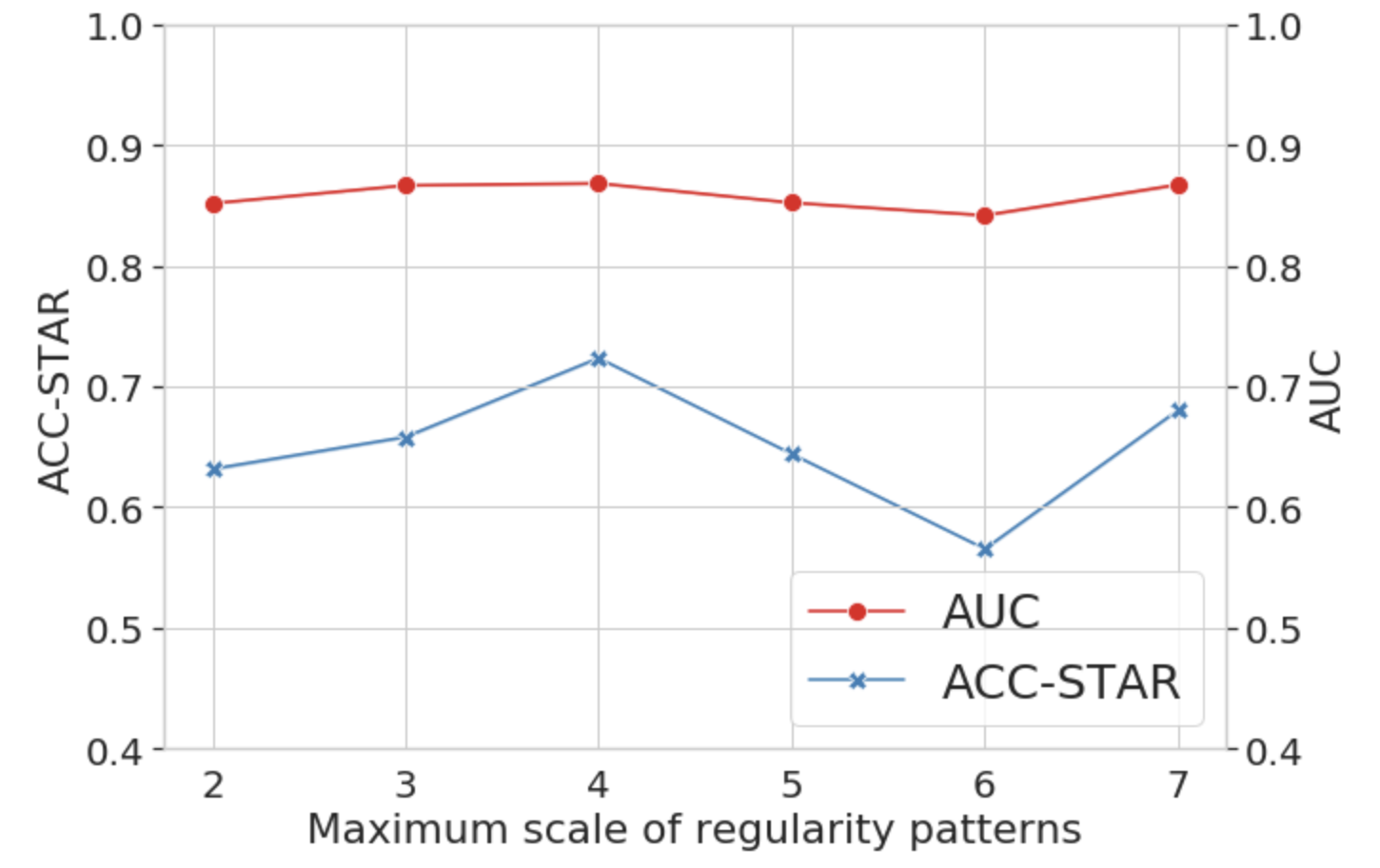}
	\caption{Results of testing the maximum scale $S$ of multi-scale bag-of-regularity.}
	\label{fig:Scale_testing}
\end{figure}
	
While we are testing the maximum scale $S$, we fix all other parameters and vary $S$ from $2$ to $7$ because the minimum time length of the repeated pattern is two days, and the course schedule is a 7-day cycle. The prediction results are shown in Fig. (\ref{fig:Scale_testing}). We found that the overall classification performance measured by AUC is not sensitive to the maximum scale $C$, but it affects a lot on the correctness of identifying STAR. EPARS achieves the best performance when $C=4$. The reason may be in two folds. One reason is that the regularity patterns of the scale 5 to 7 can be synthesized by the scale of 2 to 4. Thus it has already captured almost all regularity when setting the maximum scale $C=4$. The other reason is that the output feature vector of multi-scale bag-of-regularity is short and dense when $S=4$. It will dramatically become sparse when $S \ge 4$ in our cases, which makes the performance worse.

\begin{table}[h!]
\centering
\caption{Results of testing co-occurrence threshold $\delta$.}
\label{tab:timeinterval}
\begin{tabular}{c|c|c|c}   
\toprule
$\delta$ & Ave \#edge per week & AUC  & ACC-STAR  \\  
\hline
10 seconds & 14263 & \textbf{0.8699}    &  0.5921  \\
30 seconds & 39386 & 0.8684  & \textbf{0.7237}    \\
60 seconds & 77318 & 0.8576    & 0.6316   \\
\bottomrule
\end{tabular}
\label{Tab:test_co-occurrence_threshold}
\end{table}

\begin{table}[h!]
\centering
\caption{Results of testing linking threshold $\sigma$.}
\label{tab:threshold}
\begin{tabular}{c|c|c}   
\toprule
$\sigma$ & AUC  & ACC-STAR \\
\hline
2 times & \textbf{0.8684}   & \textbf{0.7237}    \\
3 times & 0.8615   & 0.6184 \\
4 times & 0.8554   & 0.5658 \\
5 times & 0.8122   & 0.5395 \\
\bottomrule
\end{tabular}
\label{Tab:test_linking_threshold}
\end{table}

We further test how co-occurrence threshold $\delta$ and linking threshold $\sigma$ affect the modeling of social homophily and present the results in Tab. \ref{Tab:test_co-occurrence_threshold} and \ref{Tab:test_linking_threshold}. $\delta = 30$ is the best since smaller $\delta$ will make the co-occurrence network unable to capture enough social relationship for learning the social homophily and larger $\delta$ will introduce a large number of ``familiar strangers" which also damages the prediction performance. Similar results are found in the result of testing linking threshold $\sigma$. When increase $\sigma$, both AUC and ACC-STAR are dropping. The reason is that STAR and some ordinary students go to the library less often than outstanding students so that higher $\sigma$ may filter out their social interaction and results in worse prediction performance.

\section{Conclusion}
In this paper, we propose EPARS, a novel algorithm to extract students' regularity patterns of learning and social homophily from online and offline learning behaviors for early predicting STAR.
One of our major contributions is to devise a multi-scale bag-of-regularity method to extract regularity features from sequential learning behaviors, which is robust for sparse data. In addition, we model students' social relationships by constructing a co-occurrence network from library check-in records and embed their social homophily as feature vectors.
Before training a classifier, we oversample the minority examples to overcome the label imbalance issues.
Extensive experiments are conducted on a large scale dataset covering all undergraduate students in the whole university.
Experimental results indicate that our EPARS improves the accuracy of baselines by $14.62\%\sim38.22\%$ and $5.77\%\sim34.14\%$ in predicting STAR in the first week and the last week of a semester, respectively.

\subsubsection*{Acknowledgement.}
This research has been supported by the PolyU Teaching Development (Grant No. 1.61.xx.9A5V) and ARC Discovery Project (Grant No. DP190101985, DP170103954 and DP170101172).

%
%
%
\bibliographystyle{splncs04}
\bibliography{references}
%




\end{document}